\documentclass{article}

\usepackage{arxiv}

\usepackage[utf8]{inputenc} 
\usepackage[T1]{fontenc}    
\usepackage{hyperref}       
\usepackage{url}            
\usepackage{booktabs}       
\usepackage{amsfonts}       
\usepackage{amsthm,amsmath,amssymb}
\usepackage{amsmath,amssymb}
\usepackage{bm}
\usepackage{nicefrac}       
\usepackage{microtype}      
\usepackage{lipsum}
\usepackage{graphicx}
\graphicspath{ {./images/} }
\usepackage{svg}
\usepackage{mathtools}
\usepackage{stmaryrd}
\usepackage{xcolor}

\title{Interpolating neural network: A novel unification of machine learning and interpolation theory}

\author{
 Chanwook Park \thanks{Equal contribution with Coauthor 1 and Coauthor 2}\\
 Department of Mechanical Engineering\\
 Northwestern University\\
 Evanston, IL 60208 \\
 \texttt{chanwookpark2024@u.northwestern.edu} \\
 \And
 Sourav Saha $^\ast$ \\
 Kevin T. Crofton Department of \\Aerospace and Ocean Engineering\\
 Virginia Polytechnic Institute \\and State University\\
 Blacksburg, VA 24060 \\
 \texttt{souravsaha@vt.edu} \\
 \And
 Jiachen Guo\\
 Theoretical and Applied Mechanics Program\\
 Northwestern University\\
 Evanston, IL 60208 \\
 \texttt{jiachenguo2020@u.northwestern.edu} \\
 \And
  Hantao Zhang\\
 Theoretical and Applied Mechanics Program\\
 Northwestern University\\
 Evanston, IL 60208 \\
 \texttt{hantaozhang2029@u.northwestern.edu} \\
 \And
 Xiaoyu Xie\\
 Department of Mechanical Engineering\\
 Northwestern University\\
 Evanston, IL 60208 \\
 \texttt{xiaoyuxie2020@u.northwestern.edu} \\
 \And
 Miguel A. Bessa \\
 School of Engineering\\
 Brown University\\
 Providence, RI \\
 \texttt{miguel\_bessa@brown.edu} \\
 \And
 Dong Qian \\
 Department of Mechanical Engineering\\
 University of Texas at Dallas\\
 Richardson, TX 75080 \\
 \texttt{dong.qian@utdallas.edu} \\
  \And
 Wei Chen \\
 Department of Mechanical Engineering\\
 Northwestern University\\
 Evanston, IL 60208 \\
 \texttt{weichen@northwestern.edu } \\
 \And
 Gregory J. Wagner \\
 Department of Mechanical Engineering\\
 Northwestern University\\
 Evanston, IL 60208 \\
 \texttt{gregory.wagner@northwestern.edu} \\
 \And
 Jian Cao \\
 Department of Mechanical Engineering\\
 Northwestern University\\
 Evanston, IL 60208 \\
 \texttt{jcao@northwestern.edu} \\
 \And
 Wing Kam Liu \thanks{Corresponding author}\\
 Department of Mechanical Engineering\\
 Northwestern University\\
 Evanston, IL 60208 \\
 \texttt{w-liu@northwestern.edu} \\
}

\begin{document}
\maketitle

\newpage
\begin{abstract}
Artificial intelligence (AI) has revolutionized software development, shifting from task-specific codes (Software 1.0) to neural network-based approaches (Software 2.0). However, applying this transition in engineering software presents challenges, including low surrogate model accuracy, the curse of dimensionality in inverse design, and rising complexity in physical simulations. We introduce an interpolating neural network (INN), grounded in numerical methods, to realize Engineering Software 2.0 by advancing data training, partial differential equation solving, and parameter calibration. INN offers orders of magnitude fewer trainable/solvable parameters for comparable model accuracy than traditional multi-layer perceptron (MLP) or physics-informed neural networks (PINN). Demonstrated in metal additive manufacturing, INN rapidly constructs an accurate surrogate model of Laser Powder Bed Fusion (L-PBF) heat transfer simulation, achieving sub-10-micrometer resolution for a 10 mm path in under 15 minutes on a single GPU. This makes a transformative step forward across all domains essential to engineering software.

\end{abstract}

\section{Introduction}

The evolution of software programming methodologies has transitioned from reliance on rigidly hard-coded instructions, governed by explicitly defined human rules, to the adoption of neural network-based algorithms. The transition is coined as "from \textit{Software 1.0} to \textit{Software 2.0}" \cite{Karpathy_2017}. 
The shift towards Software 2.0 partially resolved the issue of labor-intensive programming in Software 1.0 and has significantly advanced the domain of large language models (LLMs) and other foundational models \cite{moor2023foundation,vaswani2017attention}. 
However, the application of these technologies in the fields of engineering and science presents unique challenges. One such challenge is the relatively low generalization accuracy of machine learning (ML)-based partial differential equation (PDE) solvers compared to traditional numerical methods \cite{lu2021learning, zhang2022simulation, goswami2022deep, park2023convolution, grossmann2024can}. After in-depth research was conducted on those methods, an increasing number of researchers began to question their effectiveness \cite{grossmann2024can, mcgreivy2024weak}. Another challenge is the curse of dimensionality that arises when solving inverse design problems. As the design parameter space grows,  computation requirements easily become prohibitive. Furthermore, the complexity of physical simulations continues to increase. For instance, additive manufacturing (AM) or integrated circuit (IC) simulations necessitate extremely fine resolution with high accuracy and multi-physics/multi-scale analysis \cite{amin2024npj, leonor2024go, shih2021fe}. 


In this article, we propose the concept of \textit{Engineering Software 2.0}, which aims to advance the current generation of numerical and data-driven engineering software into a new paradigm.  

\begin{quote}
    \textit{\centering “Engineering Software 2.0 is an end-to-end software system that unifies and advances data training, PDE solving, and parameter calibration
    in science/engineering problems."}
\end{quote}

We introduce interpolating neural networks (INNs) to achieve the new paradigm of Engineering Software 2.0. 
INNs generalize numerical methods such as the finite element method (FEM) \cite{liu2022eighty, zhang2021hierarchical, saha2021hierarchical} as a subset of deep neural network (DNN) or graph neural network (GNN). Although different groups of researchers have developed the two (FEM and neural networks), they share one thing in common; the goal is to find a function. The former finds a solution field of a differential equation while the latter finds an input-output relationship. 
Although it is possible to formulate INN from a DNN, we adopt GNN architecture to explain INN as a special case of GNN with a set of rules (i.e., interpolation).

GNNs provide a general formulation to process information available in a graph, i.e. a cloud of interconnected points or nodes \cite{wu2020comprehensive, zhou2020graph, ju2024comprehensive}. Each node receives information from its connected neighbors that is then processed by a feedforward neural network to make a new prediction at that node. However, their network architectures are often overly complex, making convergence difficult for highly nonlinear problems. In Science and Engineering, most computer simulations also occur in interconnected point clouds (or nodes), but each interconnected node is associated with simpler interpolatory functions that are fast to differentiate and easy to integrate, leading to methods that obey established error bounds and convergence rates. This provides confidence in the predictions made but also imposes a need for relatively fine discretizations (dense point clouds) because each point is related to its connected neighbors by a simple function (e.g., low order polynomials or splines). INNs bridge the gap between the extreme generality of GNNs and the stringent restrictions imposed by numerical methods by constructing interpolatory GNNs that can simplify to known numerical methods or complexify to known GNN architectures.

\begin{figure}
    \centering
    \includegraphics[width=1.0\linewidth]{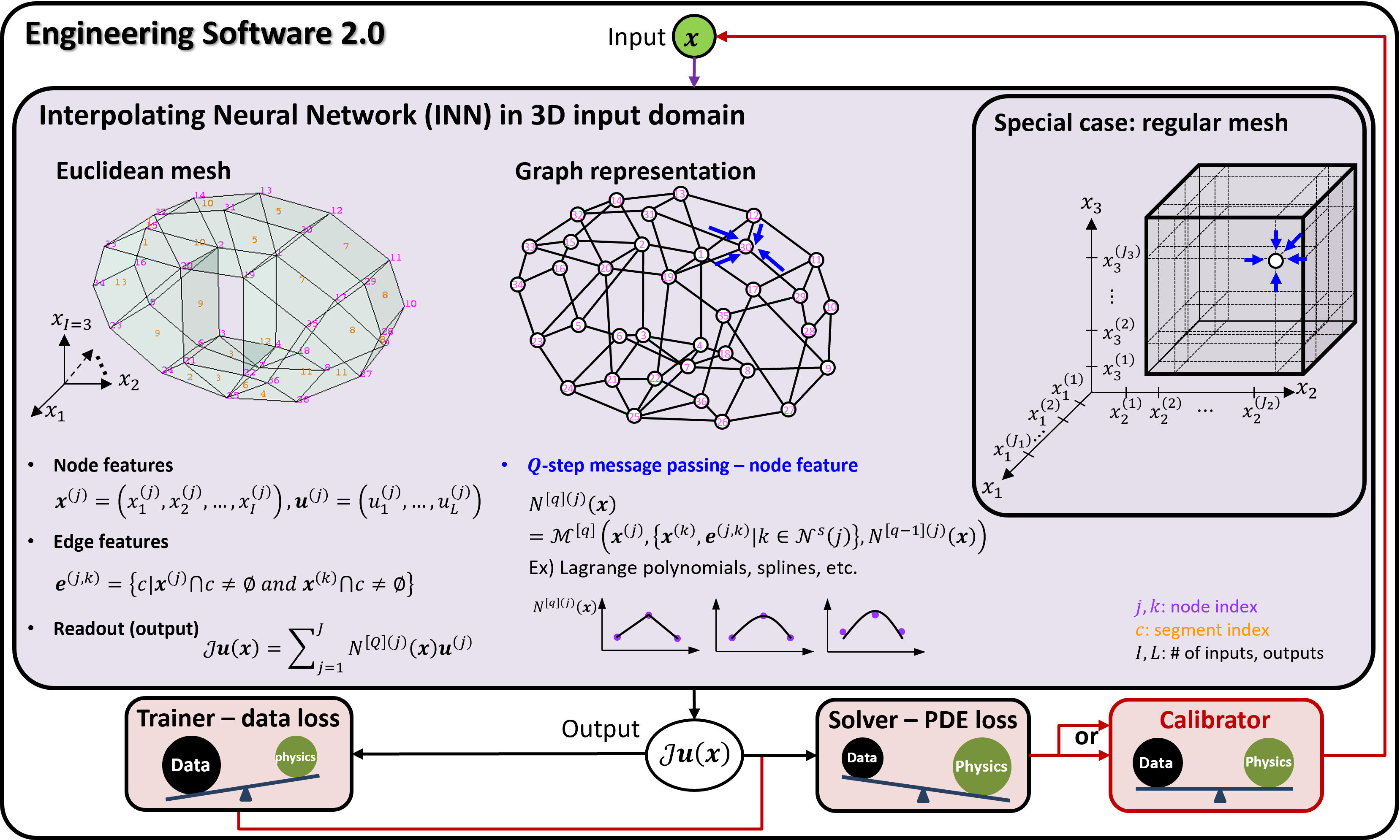}
    \caption{Overview of Engineering Software 2.0 enabled with Interpolating Neural Network (INN). The INN box illustrates the graph representation of the input domain discretized with an arbitrary Euclidean mesh (left) and a regular mesh as a special case (right). Node and edge features are given from the discretized mesh. After $Q$-step message passing, each node $j$ will store an interpolation function  $N^{[Q](j)}(\bm{x})$. Finally, the readout operation sums the product of the interpolation functions and nodal values $\bm{u}^{(j)}$. The superscripts with square brackets [] and parentheses () denote the message passing step and graph node index, respectively. The interpolation operator $\mathcal{J}$ denotes that $\mathcal{J}\bm{u}(\bm{x})$ is a function that interpolates discrete values of $\bm{u}^{(j)}$. The INN trainer employs data-driven loss functions (e.g., mean squared error loss for regression) while the INN solver adopts a residual loss of a partial differential equation (PDE). A trained/solved INN model can then be employed as a forward model of a calibrator to solve an inverse problem. 
    }
     
    \label{fig:figure1}
\end{figure}

INNs consist of three processes: 1) discretize an input domain into non-overlapping segments whose bounds are denoted by interpolation nodes, 2) construct a graph with the interpolation nodes and formulate the message passing operation as a form of interpolation functions, and 3) optimize the values and coordinates of the interpolation nodes for a given loss function. Figure \ref{fig:figure1} describes the approach. When the input domain is discretized with a regular mesh (see the special case in Figure \ref{fig:figure1}), INNs can leverage tensor decomposition (TD) \cite{kolda2009tensor, zhang2022hidenn,lu2023convolution,li2023convolution} to convert the growth of the computational cost of a high-dimensional problem from exponential to linear. 
INNs facilitate three major tasks in computational science and engineering: 1) data training, 2) PDE solving, and 3) parameter calibration, by reducing trainable parameters, training time, and memory/storage requirements without compromising model accuracy. 

A representative application of INN can be found in metal additive manufacturing (AM), where a laser is used to melt and fuse metal powder to build a metal component layer by layer. As the laser spot size ranges from 50 to 100$\mu m$ \cite{amin2024npj, attar2014selective}, physical simulations of AM require a sub-10 micron resolution that challenges most physical simulations \cite{leonor2024go, liao2023efficient}. The required computational resources become prohibitive, particularly when they are employed to generate training data for a surrogate model within a vast parametric space. As illustrated in Figure \ref{fig:figure3}(a), INN presents a new direction toward part-scale AM simulations for online manufacturing control, demonstrating performance gains by orders of magnitude. We also conduct various numerical experiments that cover computer science and engineering domains to demonstrate the superior capabilities of INNs in training, solving, and calibrating. The computer codes can be found at \href{https://github.com/hachanook/pyinn}{https://github.com/hachanook/pyinn}. 



\begin{figure}
    \centering
    \includegraphics[draft=false, width=1.0\linewidth]{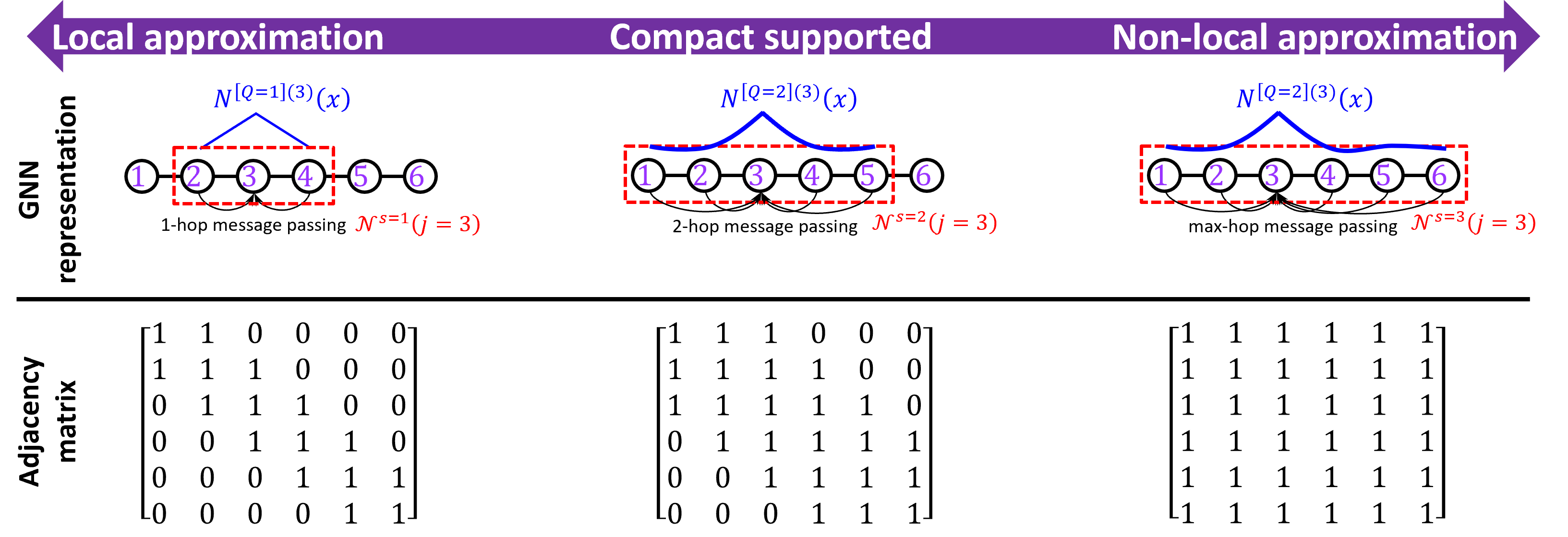}
    \caption{Illustration of a 1D input domain discretized with 5 segments and 6 nodes.}
     
    \label{fig:figure2}
\end{figure}

\section{Results}
\label{sec:INN}

Consider a regression problem that relates $I$ inputs and $L$ outputs. The first step of an INN is to discretize the input domain in the $I$-dimensional Euclidean space into a mesh, which can be as general as an unstructured irregular mesh or as specific as a structured regular mesh. In any case, a mesh in the Euclidean space can be readily represented as a graph – the most general form of a discretized input – as illustrated in the INN box of Figure \ref{fig:figure1}. The conversion from mesh to graph has been widely accepted in the literature \cite{alet2019graph, michelis2022physics, gao2022physics, xue2022physics}. Suppose there are $J$ nodes (or vertices) and $E$ edges in the graph, where the input domain is discretized with $C$ non-overlapping segments. Each segment occupies a subspace of the $I$-dimensional input domain. 

Each node (e.g., node $j$) has features: nodal coordinates $(\bm{x}^{(j)}\in\mathbb{R}^I )$ and values $(\bm{u}^{(j)}\in \mathbb{R}^L )$. Note that the superscript with parentheses refers to the graph node index. An edge that links node $j$ and $k$ has a feature $\bm{e}^{(j,k)}$ that stores indices of segments connected to the edge. For example, when $I=3, L=2$, node $10$ in Figure \ref{fig:figure1} stores $\bm{x}^{(10)}=(x^{(10)}_1,x^{(10)}_2,x^{(10)}_3)$ and $\bm{u}^{(10)}=(u^{(10)}_1,u^{(10)}_2)$. The edge connecting nodes $30$ and $12$ stores $\bm{e}^{(30,12)} = \{5,7\}$.

A typical message passing in a graph neural network (GNN) returns a hidden state for each graph node which is mostly a tensor (including matrix and vector) \cite{wu2020comprehensive}. In contrast, the INN message passing returns a function (i.e., interpolation function $N^{(j)}(\bm{x})$, or often called \textit{shape function} in FEM) for each node $(j=1, \dots, J)$ as a hidden state. A general $Q$-step message passing of an INN, $\mathcal{M}^{[q]}(*)$, can be expressed as:

\begin{equation} \label{eq:p-step_message_passing}
    N^{[q](j)}(\bm{x}) = \mathcal{M}^{[q]} \left( \bm{x}^{(j)}, \{ \bm{x}^{(k)}, \bm{e}^{(j,k)} | k\in \mathcal{N}^s(j) \}, N^{[q-1](j)}(\bm{x})  \right), \quad q=1,\cdots,Q, \quad N^{[0](j)}(\bm{x})=0,
\end{equation}

where $N^{[q](j)}(\bm{x})$ is the interpolation function at node $j$ after $q$-th message passing and $\mathcal{N}^s(j)$ is a set of neighboring nodes of the center node $j$ with $s$ connections (i.e., $s$-hops, see Figure \ref{fig:figure2} and Supplementary Information (SI) Section 1.1 for visual illustrations). It is worth noting that the interpolation functions satisfy the Kronecker delta property, i.e., $N^{[q](j)}(\bm{x}^{(k)})=\delta_{jk}$. The operation $\mathcal{M}^{[q]}(*)$ constructs an interpolation function $N^{[q](j)}(\bm{x})$ for a graph node $\bm{x}^{(j)}$ using neighboring nodal coordinates and edge information: ${\bm{x}^{(k)}, \bm{e}^{(j,k)}}$, and the interpolation function of the previous message passing: $N^{[q-1](j)}(\bm{x})$.


When there is only one message passing ($Q=1$) with $s=1$ hop, INNs degenerate to the standard FEM linear shape function (or piecewise linear interpolation). As illustrated in Figure \ref{fig:figure2} (left), FEM with linear elements is the most localized approximation of an INN. On the other hand, one can progressively enlarge the support domain and the approximation capability of an interpolation function by adding message passings with a higher hop. For instance, the second message passing with $s=2$ hop constructs compact-supported interpolation functions with higher nonlinearity. This is theoretically equivalent to the generalized finite element method (GFEM) \cite{tian2013extra} and convolution hierarchical deep learning neural networks (C-HiDeNN) \cite{park2023convolution, lu2023convolution}. The message passing with a max-hop (Figure \ref{fig:figure2}, right) can even make it a non-local approximation where support of the interpolation functions occupy the global domain, mimicking meshfree methods that exhibit superconvergence (i.e., faster convergence rate than the complete order of polynomial basis) \cite{liu1995reproducing, liu1995reproducing2, leng2019super}. Depending on the choice of interpolation technique, other hyper-parameters can be involved in $\mathcal{M}^{[q]}(*)$ such as the activation (or basis) function and dilation parameter \cite{park2023convolution, lu2023convolution}. See SI Section 1 for various choices of the message passing operation.

Regardless of the choice of interpolation technique for the message passing, the global degrees of freedom (i.e., number of nodes) remain constant. This distinguishes INNs from higher-order FEM where each segment is enriched with additional nodes. INNs, on the other hand, adapt nodal connectivity through the adjacency matrix and basis functions to reproduce almost any interpolation technique available in numerical methods. The compact-supported interpolation functions enable INNs to be optimized locally, making INNs distinguishable from MLPs. Since an MLP is a global approximator, INNs with compact-supported interpolation functions 1) converge faster than MLPs with the same number of trainable parameters (see Section \ref{subsec:benchmark_INN_trainer}) and 2) facilitate training on a sparse dataset (see Section \ref{subsec:applications_calibration}), although it might end up with overfitting. 


During the forward propagation, an input variable $\bm{x} \in \mathbb{R}^I$ enters each graph node’s interpolation function $N^{[Q](j)}(\bm{x})$, followed by a graph-level readout operation: 

\begin{equation} \label{eq:readout_general}
    \mathcal{J}\bm{u}(\bm{x}) = \sum^{J}_{j=1}{N^{(j)}(\bm{x}) \bm{u}^{(j)}}, \quad \bm{x} \in \mathbb{R}^I, \bm{u} \in \mathbb{R}^L,
\end{equation}

where the superscript $[Q]$ is dropped for brevity (i.e., $ N^{[Q](j)}(\bm{x}) = N^{(j)}(\bm{x})$). The interpolation operator $\mathcal{J}$ will designate an interpolated field output throughout this paper. The readout operation can be written as a tensor contraction (or matrix multiplication):

\begin{equation} \label{eq:INN_forward_propagation}
    \mathcal{J}\bm{u}(\bm{x}) = \begin{bmatrix} | & | & | & | \\ \bm{u}^{(1)} & \bm{u}^{(2)} & \cdots & \bm{u}^{(J)} \\ | & | & | & |  \end{bmatrix} \cdot \begin{bmatrix} N^{(1)}(\bm{x}) \\ N^{(2)}(\bm{x}) \\ \vdots \\ N^{(J)}(\bm{x})  \end{bmatrix} = \bm{U}\bm{\mathcal{X}(x)},
\end{equation}

where $\bm{u}^{(j)} \in \mathbb{R}^L, \bm{U} \in \mathbb{R}^{L\times J}, \bm{\mathcal{X}(x)} \in \mathbb{R}^{J}$. The matrix $\bm{U}$ is a stack of nodal values $\bm{u}^{(j)}$, while $\bm{\mathcal{X}(x)}$ is a vectorized function of $\bm{x}$ that is parameterized with nodal coordinates $\bm{x}^{(j)}$ in the message passing operation.

The graph node features (i.e., coordinate ($\bm{x}^{(j)}$) and value ($\bm{u}^{(j)}$)) are trainable parameters of the INN. If the nodal coordinates ($\bm{x}^{(j)}$) are fixed, one can find nodal values ($\bm{u}^{(j)}$) without changing the discretization of the input domain. If the nodal coordinates are also updated, the optimization will adjust the domain discretization similar to r-adaptivity in FEM \cite{zhang2021hierarchical, saha2021hierarchical}. Once the forward propagation is defined, the loss function is chosen based on the problem type: training, solving, or calibrating (see Section \ref{sec:methods} for details). 

When the input domain is discretized with a regular mesh, as illustrated in the special case of Figure \ref{fig:figure1}, we can significantly reduce the trainable parameters (or degrees of freedom, DoFs) by leveraging Tensor Decomposition (TD) \cite{kolda2009tensor}. Section \ref{sec:methods} introduces the two widely accepted TD methods: Tucker decomposition \cite{tucker1963implications, tucker1966some} and CANDECOMP/PARAFAC (CP) decomposition \cite{harshman1970foundations, carroll1970analysis, kiers2000towards}.

\begin{figure}
    \centering
    \includegraphics[draft=false, width=1.0\linewidth]{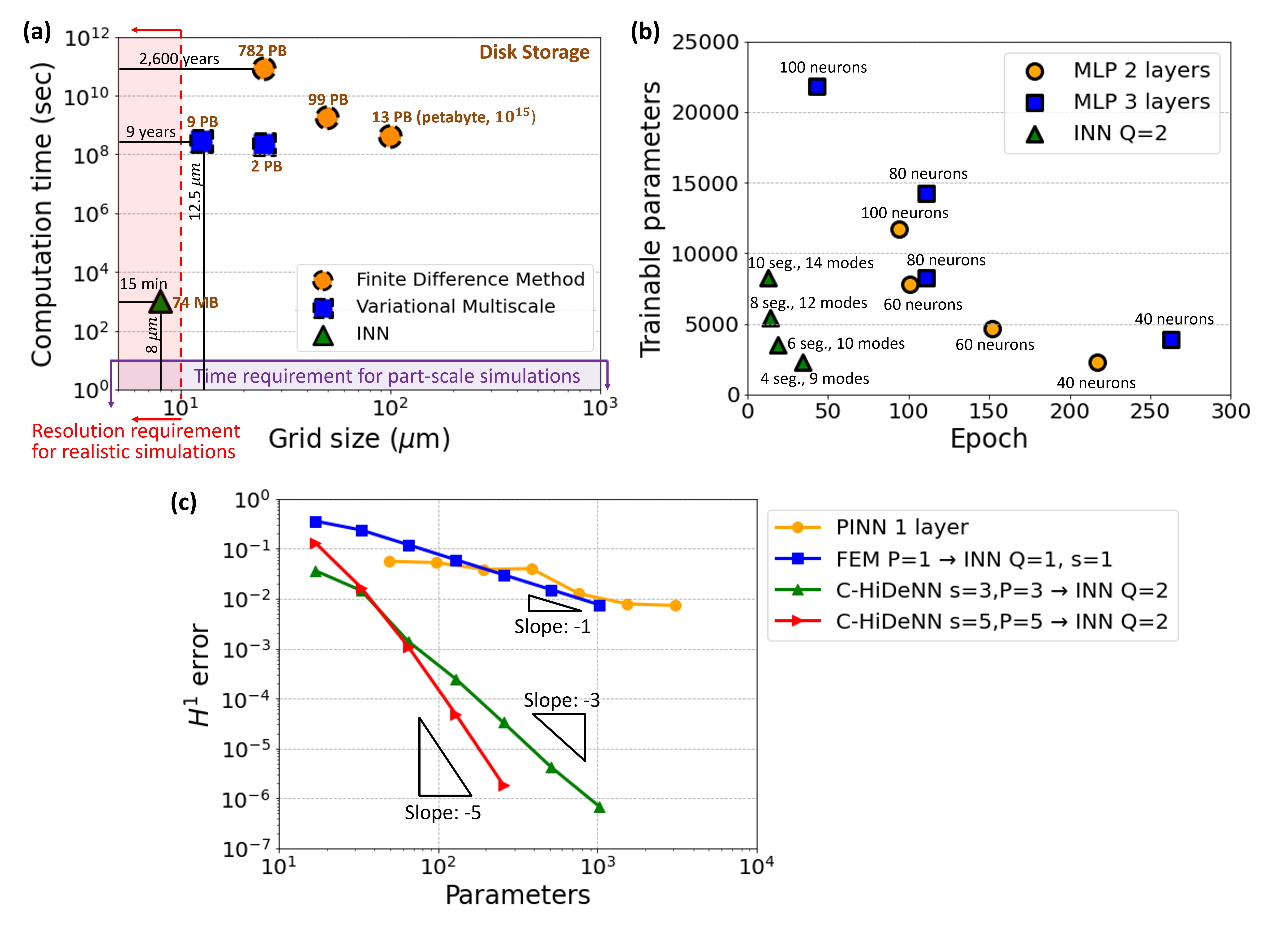}
    \caption{Benchmarks for the INN solver (a) and the INN trainer (b). In (a), a 3D-space, time, and 4D-parameter heat transfer equation is solved to model a 10 mm single-track laser powder bed fusion (L-PBF) metal AM. Detailed problem definition and explanation can be found in Section \ref{sec:applications}. We compare the single-scale finite difference method (FDM) solver \cite{liao2023efficient}, the variational multiscale FEM solver \cite{leonor2024go}, and the INN solver with CP decomposition and Q=2. The data points for the first two methods with dashed marker edges are estimated, while that of INN with a solid marker edge is computed. In (b), a standard data training (regression) problem is studied. Each trainer stops when the training loss hits the stopping criteria (MSE:4e-4).  We set the same optimization condition: ADAM optimizer; learning rate: 1e-3; batch size: 128. The number of neurons per hidden layer is denoted for each MLP data point while the number of segments and modes are denoted for each INN data point. INN with CP decomposition and Q=2 is used. In (c), a 1D Poisson's equation is solved with PINN and INN. PINN is made of a 1-layer MLP with a varying number of neurons. Randomly selected 10k collocation points are used to compute the PDE loss. With a batch size of 128, 10k epochs are trained using an ADAM optimizer with a learning rate of 1e-1. FEM with linear elements is equivalent to an INN with Q=1, s=1. C-HiDeNNs with s=3,P=3 and with s=5,P=5 are equivalent to INNs with Q=2, s=3, P=3, and with Q=2, s=5, P=5, respectively. All solvers and trainers are graphics processing units (GPU) optimized with the JAX library \cite{jax2018github}.
    }
     
    \label{fig:figure3}
\end{figure}

\subsection{Benchmarking INN trainer}
\label{subsec:benchmark_INN_trainer}

To highlight INN's advantages in speed (training epoch) and storage (number of parameters) over traditional MLP, we introduce a benchmark problem: a 10-input 5-output physical function \cite{surjanovic2013virtual}, (see SI Section 2.4 for details). Using this deterministic equation, we randomly generate 100,000 data using a Latin hypercube sampling. The dataset is split into 80\% and 20\% for train and test, respectively. MLPs with two and three hidden layers and with a sigmoid activation are tested while INNs with CP decomposition, two (Q=2) message passing, and $s=2,P=2$ polynomial activation are adopted. To investigate the convergence behavior depending on the number of parameters, we set the stopping criteria as training MSE: 4e-4 and count the epoch at convergence. Since the physical function is deterministic (no noise) and we drew a sufficiently large number of data, no considerable overfitting was observed.

Figure \ref{fig:figure3}(b) reveals that given the same number of trainable parameters, INNs converge significantly faster than MLPs. For instance, the INN with 2,250 parameters converges at the 34th epoch while the MLP with 2,285 parameters converges at the 217th epoch. MLPs with more parameters tend to stop at earlier epochs, however, even the largest MLP with 21,805 parameters converges at the 45th epoch while that of INN with only 8,250 parameters converges at the 13th epoch. This benchmark demonstrates that the INN trainer is lightweight and fast-converging compared to traditional MLPs. Other benchmarks of the INN trainer are elucidated in SI Section 2.

\subsection{Benchmarking INN solver}
\label{subsec:Benchmark_INN_solver}

INNs can achieve a proven convergence rate when solving a PDE. In numerical analysis, the convergence rate is the decaying rate of an error measurement as the mesh refines. This benchmark solves a 1D Poisson's equation defined in Section \ref{subsec:Poisson_equation} with the $H^1$ norm error estimator. It is mathematically proven that a numerical solution of FEM or any interpolation-based solution method exhibits a convergence rate of $P$, which is the complete order of polynomial basis used in the interpolation \cite{lu2023convolution}.

Figure \ref{fig:figure3}(c) is the $H^1$ error vs. parameters (i.e., degrees of freedom) plot where the convergence rates are denoted as the slope of the log-log plot. Depending on the message passing operation, INNs degenerate to linear FEM ($P=1$) or C-HiDeNNs (or GFEM) with different order of polynomials: $P=3, P=5$. However, PINNs constructed with MLP do not reveal a convergent behavior. Only a few theoretical works have proved the convergence rate of PINNs under specific conditions \cite{shin2020convergence}. A PDE solver needs to have a stable and fast convergence rate because it guides an engineer in choosing the mesh resolution and other hyperparameters for achieving the desired level of accuracy. We also demonstrate the convergent behavior of INN solvers with and without CP decomposition for a 3D linear elasticity equation in SI Sections 4.5 and 4.6.


\section{Discussion}
\label{sec:applications}

\begin{figure}
    \centering
    \includegraphics[width=1.0\linewidth]{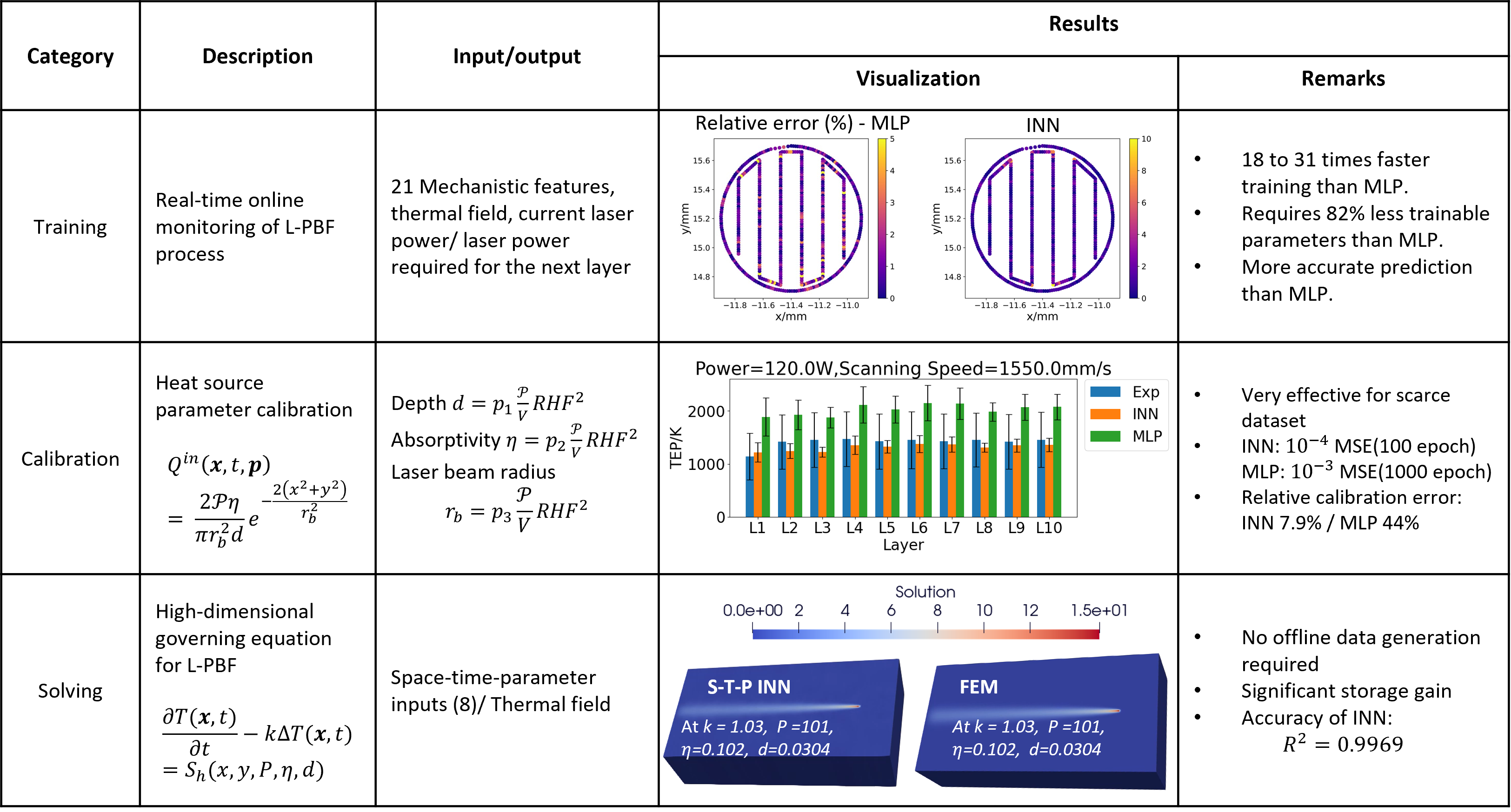}
    \caption{A summary of engineering applications of INNs to additive manufacturing (AM) problems. First, a data-driven real-time online monitoring and feedback control tool is formulated with an INN that uses only 18\% training parameters of MLP and is 18-31 times faster than MLP. The calibration problem develops a reduced-order model of laser powder bed fusion (L-PBF) AM to calibrate the heat source parameters from experimental data. Finally, the INN solver solves a space-time-parameter heat transfer equation, resulting in a significant storage reduction and faster simulations. }
    \label{fig:figure4}
\end{figure}

Now we illustrate how Engineering Software 2.0 enabled with INNs can be applied to advance various facets of metal additive manufacturing (AM). Note that the INN with CP decomposition will be referred to as INN in this section for brevity. Laser powder bed fusion (L-PBF) is a type of metal AM in which metal powders are layered and melted by a laser beam to create objects that match the provided CAD designs \cite{bhavar2017review}. The added flexibility of L-PBF comes with a huge design space and spatially varying microstructure after manufacturing. Therefore, considerable research is devoted to employing in situ monitoring data for real-time control of L-PBF processes to minimize the variability and optimize the target properties \cite{cai2023review}. Computational modeling of L-PBF spans the simulation of the manufacturing process with a high dimensional design space (solving), the identification of model parameters from sparse experimental data (calibrating), and the online monitoring and control of the L-PBF (training). 

The governing equation for modeling the manufacturing process is given by a heat conduction equation with a moving laser heat source $S_{h}$:

\begin{equation}
\frac{\partial T(\mathbf{x},t)}{\partial t}-k\Delta T(\mathbf{x},t)= S_{h}(\mathbf{x},t,P,\eta,d)  
\label{eq:AM}
\end{equation}

where the equation involves spatial variables $\mathbf{x}$, temporal variable $t$, and four variable parameters: thermal conductivity $k$, laser power $P$, absorptivity $\eta$, and laser depth $d$. For the simplest case of single track laser pass, the initial condition is $T(\mathbf{x},0)=T_0(\mathbf{x})$, and the boundary conditions (BCs) are $T=\Tilde{T}$ in $\Gamma^D$ for Dirichlet BC and $\frac{\partial T}{\partial \mathbf{x}}\cdot \mathbf{n}=\Bar{q}$ in $\Gamma^N$ for Neumann BC, where the spatial domain boundary $\Gamma = \Gamma^D \cup \Gamma^N$.

\subsection{Real-time online control of laser powder bed fusion additive manufacturing}
\label{subsec:applications_training}

The first application is fully data-driven, where the data are generated from numerical methods and the INN trainer is used to develop an online control system for L-PBF AM. The goal is to maintain a homogeneous temperature of the melt pool (i.e., a molten region near the laser spot) across each layer using real-time sensor data, under the hypothesis that \textit{a homogeneous melt pool temperature across each layer of material deposition will reduce the variability in the microstructure and mechanical properties}. The laser power at each point of a certain layer needs to be controlled (increased/decreased by a certain amount compared to the previous layer) to achieve this homogeneity in melt pool temperature. The fundamental challenge of implementing a model predictive control system for such an application is the computational resources required for forward prediction (involving finite element or computational fluid dynamics models) and inverse prediction, as illustrated in Figure \ref{fig:figure4}, the row for training. The INN trainer aims to provide a reliable and memory-efficient surrogate model that can replace mechanistic models in predictive control systems. In this application, the database is generated by fusing experiments and computational methods (see SI Section 2.6 for details) for aluminum alloy. The online control loop uses the trained INN models to inform the manufacturing machine. There are two models: \textit{forward model} and \textit{inverse model}. 

The forward model uses the mechanistic features and the thermal emission plank (TEP, a representative of the melt pool temperature) to predict the TEP of the next layer of the build when no correction is applied. The mechanistic features are precomputed based on the tool path and geometry of the build to generalize a data-driven model for unseen tool path and geometric features \cite{kozjek2022data}). The target TEP for the next layer is the average of the forward model's outputs (i.e., predicted TEPs for the next layer) measured at the current layer. The inverse model takes this target TEP as the input along with the previous layer's mechanistic features to predict the required laser power for the next layer. Finally, the control software calculates how much change in laser power is required at each point, and sends the information to the machine. The corrected laser power is then used to print the next layer. 

The forward model is trained on experimentally observed data, while the inverse model is trained on computational data coming from a finite-difference solver \cite{liao2023efficient}. The INN model is applied for both forward and inverse prediction and is compared with the MLP. Figure \ref{fig:figure4} shows that for the inverse model INN is at least 18 to 31 times faster to train compared to the MLP to reach the same level of training error, and it reduced the number of trainable parameters by a staggering amount of 82\%. The speed-up depends on the amount of data used. The INN method is consistently more accurate compared to MLP (see, 
SI Section 2.6 for more details).

\subsection{Calibration of heat source parameters for additive manufacturing}
\label{subsec:applications_calibration}

In this example, an INN calibrates the heat source function $S_h$ in Eq.\ref{eq:AM} for experimental data. A Gaussian beam profile is modeled as a volumetric heat source \cite{li2024statistical}, which is written as:

\begin{equation} \label{eq:heatsource}
S_h (\mathbf{x},t,\mathbf{p}) = \frac{2P\eta}{\pi r^{2}_{b}d}e^{-\frac{2(x^2+y^2)}{r^{2}_{b}}}\cdot \mathbf{1}_{\{z|z_{top}-z\leq d\}}(z)
\end{equation}

where $z_{top}$ is the z-coordinate of the current surface of the powder bed and $\mathbf{1}$ is the indicator function that returns 1 if the condition is met and 0 otherwise. The notation follows Eq. \ref{eq:AM}, and $r_b$ is the Gaussian profile's standard deviation that controls the beam's width. Following reference \cite{amin2024npj}, the heat source parameters $d, \eta, r_b$ are controlled by the calibration parameters $p_1$, $p_2$, $p_3$ such that $d = p_1\frac{P}{V}RHF^2$, $\eta = p_2\frac{P}{V}RHF^2$, and $r_b=p_3\frac{P}{V}RHF^2$. Here, $V$ is the laser scan speed and $RHF$ is a predetermined residual heat factor based on the laser toolpath \cite{yeung2020residual}. 

The major challenge of this problem is to achieve a highly accurate surrogate model given a sparse set of training data. INNs outperform MLPs in training these scarce data because they are rooted in the interpolation theory. As a demonstration, an INN with 1760 trainable parameters is trained on 28 sets of meltpool temperature data generated by the finite difference method (FDM) \cite{liao2023efficient} using 4 different power-scanning speed set-ups. Later, this trained INN is used as a surrogate to optimize the three parameters ($p_1,p_2,p_3$) given the experimental data. For comparison, a 3-layer MLP with 2689 trainable parameters (roughly the same order of magnitude as the INN) is trained using the same data. We observed that INNs converge well within 100 epochs, while MLPs require over 1000 epochs to converge to a test MSE 10 times larger than the INN given all other setups the same. Details on this training and calibration process are provided in SI Section 3.1.

The INN- (definition of the INN calibrator can be found in Section \ref{subsec:INN_calibrator_loss} and MLP-based calibrations are repeated 50 times to provide a statistical comparison (see Figure \ref{fig:figure4}). Even with only 28 sets of simulation data, the calibrated parameters using the INN produce a mean melt pool temperature within a 7.9\% difference of the experimental data. In contrast, the parameters calibrated from the MLP surrogate produce a difference of 44. 1\%, which is unsatisfactory. This is expected since MLP is a general model requiring a large amount of data to achieve a reliable representation of the problem. On the other hand, the INN achieves a much lower difference, showing that the model can learn underlying physics even with a scarce dataset.

\subsection{Solving high dimensional Space-Time-Parameter (S-T-P) heat equation}
\label{subsec:applications_solving}

INNs can also be utilized as a solver to directly obtain the surrogate model of the additive manufacturing process involving Space-Time-Parameter (S-T-P) dependencies without generating training data. In this case, an INN is used as a parametric interpolation function in the S-T-P solution space that satisfies the governing physical equation and corresponding boundary/initial conditions. Here, we showcase the power of INN by solving a single-track scan in the L-PBF process. The domain size is 10mm$\times$5mm$\times$2.5mm with a mesh resolution of 8$\mu$m. The laser scan speed is 250mm/s. We are interested in obtaining the temperature field which not only depends on space $\mathbf{x}$ and time $t$ but also on the laser and material parameters $\mathbf{p}$. In other words, the surrogate model $f$ is a mapping from the S-T-P continuum to the temperature field: $f:=\mathbb{R}^8 \rightarrow \mathbb{R}$.

The surrogate model $f$ can also be obtained using the standard data-driven approach. In the offline stage, training data are generated by repeatedly running numerical solvers (such as FDM or variational multiscale method (VMS)) in the Space-Time domain with different sets of parameters $\mathbf{{p}}$. Then a surrogate model is trained with the data. The data-driven approach suffers from the curse of dimensionality when the parametric space is high-dimensional, resulting in expensive computation for repetitive data generation, huge memory costs for running full-scale simulations, and excessive disk storage of offline training data. We estimate the time and storage required for the data-driven approaches with FDM and VMS, as shown in Table \ref{tab:perf} and illustrated in Figure \ref{fig:figure3}(a). 

Unlike the data-driven approaches, INN solvers treat the parameters $\mathbf{p}$ as additional parametric inputs. As a result, INNs obtain a parametric surrogate model  $f = u(\mathbf{x}, t, \mathbf{p})$ directly from the governing equation without going through the cumbersome offline data generation and surrogate model training. Moreover, our results prove the INN solution (or the S-T-P surrogate model) achieves an exceptionally high accuracy of $R^2$ = 0.9969, considering that most data-driven approaches for S-T-P problems trained on numerical simulation data suffer from low model accuracy, typically below $R^2=0.9$ \cite{karkaria2024towards}. It is worth mentioning that the physics-informed neural networks (PINNs) can handle the S-T-P problem similar to the INN solver. However, the number of collocation points scales exponentially with the number of inputs, and our preliminary study revealed that PINN fails to converge for the same 8-dimensional S-T-P problem after consuming considerable computational resources (see SI Section 4.4).
The performance comparison of INN versus data-driven methods is shown in Table. \ref{tab:perf}.

\begin{table}[]
\caption{Performance comparison of different surrogate models. The INN's discretization in the 4D parametric space is $100^4$, so the corresponding sampling points required for the data-driven methods should be $10^8$. The total simulation time and storage requirement of data-driven approaches are estimated from a single run. Detailed explanations for this benchmark can be found in SI Sections 4.3 and 4.4.}
\label{tab:perf}
\begin{center}
\begin{tabular}{cccccc}
\hline
Method & \begin{tabular}[c]{@{}c@{}}Resolution\\  ($\mu$m)\end{tabular} & \begin{tabular}[c]{@{}c@{}}Number of \\ offline simulations\end{tabular} & \begin{tabular}[c]{@{}c@{}}Total offline \\ simulation time \end{tabular} & \begin{tabular}[c]{@{}c@{}}Storage\\  (GB)\end{tabular} \\ \hline
Data-driven (FDM) & 25  & $1\times 10^8$ & 2,600 years & $7.82\times10^{8}$ \\
Data-driven (VMS) & 12.5 & $1\times 10^8$ & 9 years & $8.79\times10^6$ \\
INN & 8  & $1$ & 15 minutes & $7.37\times 10^{-2}$ \\ \hline
\end{tabular}
\end{center}
\end{table}

\section{Conclusion}
\label{sec:Conclusion}

This article demonstrates that interpolating neural networks (INNs) can train, solve, and calibrate scientific and engineering problems that are extremely challenging or prohibitive for existing numerical methods and machine learning models. The keys to INN's success are 1) it interpolates the graph nodes' values with well-established interpolation theories and 2) it leverages the tensor decomposition (TD) to resolve the curse of dimensionality. Due to the reduced number of parameters without compromising predictive accuracy, INNs become an efficient substitute for MLPs or PINNs.

There are numerous research questions that could generate significant interest across various fields, including, but not limited to, scientific machine learning, applied mathematics, computer vision, and data science. To mention a few:

\begin{itemize}
  \item Depending on the problem, INN's superior training efficiency may overfit data. The ensemble training (i.e., training on random subsets of the training data) might be a way to enhance the generality of the INN model for real-world data.
  \item Multi-resolution aspect (i.e., a varying mesh resolution across TD modes) of INNs needs to be investigated. See SI 1.4 for further discussions.
  \item The interpretability of INNs should be investigated. INNs may address the challenge of model-based interpretability (a terminology defined from \cite{murdoch2019definitions}), which involves developing models that are both simple enough to be easily understood by developers and capable of maintaining high predictive accuracy.
  \item Similar to INN solvers, the convergence of INN trainers can be studied both mathematically and numerically. 
  \item Since INNs can be used as both a solver for physics-based problems and a function approximator for data-driven problems, it is of interest to combine the two behaviors into one single model to solve large-scale problems involving both complex physics and scarce data.
  \item The superior performance of INN solvers may facilitate multiscale analysis in a vast parametric space. One can integrate the S-T-P INN solver with a concurrent multiscale analysis framework such as self-consistent clustering analysis \cite{liu2016self, yu2019self} and open a new direction towards parameterized multiscale analysis.
  \item The current INN code has been optimized to run on a single GPU. Since the INN forward pass over multiple modes can be parallelized, multi-GPU programming will further speed up the code.
  \item Although Tucker decomposition is introduced in Section \ref{subsec:Tucker_Decomposition}, the numerical experiments conducted in this article mostly focus on the CP decomposition. Theoretically, Tucker decomposition has a larger approximation space than CP decomposition. Further studies on the numerical aspects of the Tucker decomposition are needed.

\end{itemize}

We expect INNs will open a next-generation Engineering Software 2.0 that revolutionizes almost all fields of computational science and engineering.

\section{Methods}
\label{sec:methods}

\subsection{Special case 1: regular mesh with Tucker decomposition}
\label{subsec:Tucker_Decomposition}

One of the widely used TD methods is Tucker decomposition \cite{tucker1963implications, tucker1966some}. It approximates a high-order tensor with $J_i$ nodes in $i$-th dimension as a tensor contraction between dimension-wise matrices and a core tensor $\bm{\mathcal{G}}$, which has the same order of the original tensor but with smaller nodes $M_i$ ($< J_i$). The $M_i$ is often called a "mode" to distinguish it from the original node $J_i$.

Consider a three-input ($I=3$) and one output ($L=1$) system, and assume the input domain is discretized with $J=J_1 \times J_2 \times J_3$ nodes, as shown in the special case box of Figure \ref{fig:figure1}. To facilitate tensor notation, the nodal values will be denoted with left/right super/sub scripts, $\prescript{m}{i}{u}^{(j)}_l$, where $i \in \mathbb{N}^I$ is the input index, $m \in \mathbb{N}^{M_i}$ is the mode index, $l \in \mathbb{N}^L$ is the output index, and $j \in \mathbb{N}^{J_i}$ is the nodal index.

The interpolated field $\mathcal{J}u(\bm{x}) \in \mathbb{R}^{L=1} $ can be represented as a Tucker product:

\begin{equation} \label{eq:tucker_decomposition}
    \mathcal{J}u(\bm{x}) 
    = \llbracket \bm{\mathcal{G}} ; (\mathcal{J}\prescript{}{1}{\bm{u}})^T, (\mathcal{J}\prescript{}{2}{\bm{u}})^T, (\mathcal{J}\prescript{}{3}{\bm{u}})^T \rrbracket  = \bm{\mathcal{G}} \times ^2_1 (\mathcal{J}\prescript{}{1}{\bm{u}})^T \times ^2_2 (\mathcal{J}\prescript{}{2}{\bm{u}})^T \times ^2_3 (\mathcal{J}\prescript{}{3}{\bm{u}})^T ,
\end{equation}

where the core tensor $\bm{\mathcal{G}} \in \mathbb{R}^{M_1\times M_2\times M_3}$ is a trainable full matrix typically smaller than the original tensor that compresses the data. Here, $\bm{\mathcal{A}}\times^b_a \bm{\mathcal{B}}$ denotes the tensor contraction operation between the $a$-th dimension of tensor $\bm{\mathcal{A}}$ and the $b$-th dimension of tensor $\bm{\mathcal{B}}$. The $\mathcal{J}\prescript{}{i}{\bm{u}} (x_i) \in \mathbb{R}^{M_i \times 1} $ is one-dimensional (1D) interpolated output of $i$-th input dimension over $M_i$ modes represented as:

\begin{equation} \label{eq:interpolation}
    \mathcal{J}\prescript{}{i}{\bm{u}} (x_i) = \begin{bmatrix} \prescript{1}{i}{u}^{(1)} & \prescript{1}{i}{u}^{(2)} & \cdots & \prescript{1}{i}{u}^{(J_i)} \\ \prescript{2}{i}{u}^{(1)} & \prescript{2}{i}{u}^{(2)} & \cdots & \prescript{2}{i}{u}^{(J_i)} \\ \vdots & \vdots & \vdots & \vdots \\ \prescript{M_i}{i}{u}^{(1)} & \prescript{M_i}{i}{u}^{(2)} & \cdots & \prescript{M_i}{i}{u}^{(J_i)} \end{bmatrix} \cdot \begin{bmatrix} N^{(1)}(x_i) \\ N^{(2)}(x_i) \\ \vdots \\ N^{(J_i)}(x_i)  \end{bmatrix} = \prescript{}{i}{\bm{U}} \times^1_2 \prescript{}{i}{\mathcal{X}(x_i)} = \prescript{}{i}{\bm{U}}\prescript{}{i}{\mathcal{X}(x_i)},
\end{equation}

where $\prescript{}{i}{\bm{U}} \in \mathbb{R}^{M_i \times J_i}$ and $\prescript{}{i}{\bm{\mathcal{X}(x)}} \in \mathbb{R}^{J_i}$. As illustrated in the special case box of Figure \ref{fig:figure1}, the message passing (blue arrow) only happens in the axial directions, yielding 1D interpolation functions: $N^{(J_i)}(x_i)$.

When there are more than one output $(L>1)$, the interpolated field becomes a vector of $L$ elements:

\begin{equation} \label{eq:tucker_multi_output}
    \begin{aligned}
    \mathcal{J}\bm{u}(\bm{x})  &= \left[ \mathcal{J}u_1(\bm{x}), \mathcal{J}u_2(\bm{x}), \cdots \mathcal{J}u_L(\bm{x}) \right] , \textit{where} \\
    \mathcal{J}u_l(\bm{x}) 
    &= \llbracket \bm{\mathcal{G}}_l ; (\mathcal{J}\prescript{}{1}{\bm{u}}_l)^T, (\mathcal{J}\prescript{}{2}{\bm{u}}_l)^T, (\mathcal{J}\prescript{}{3}{\bm{u}}_l)^T \rrbracket  = \bm{\mathcal{G}}_l \times ^2_1 (\mathcal{J}\prescript{}{1}{\bm{u}}_l)^T \times ^2_2 (\mathcal{J}\prescript{}{2}{\bm{u}}_l)^T \times ^2_3 (\mathcal{J}\prescript{}{3}{\bm{u}}_l)^T ,
    \end{aligned}
\end{equation}
where $\mathcal{J}\prescript{}{i}{\bm{u}}_l \in \mathbb{R}^{M_i \times 1}$. The trainable parameters are the core tensors $\mathcal{G}_l \in \mathbb{R}^{M_1 \times M_2 \times M_3}, l = 1, \dots, L$, and the nodal values $\prescript{}{i}{\bm{U}}_l \in \mathbb{R}^{M_i \times J_i}, l = 1, \dots, L$, yielding a total count of $L \left( \prod^{I}_{i}{M_i} + \sum^{I}_{i}{M_i J_i} \right) $ that scales linearly with the nodal discretization $J_i$.

\subsection{Special case 2: regular mesh with CP decomposition}
\label{subsec:CP_decomposition}

Tucker decomposition can be further simplified to CANDECOMP/PARAFAC (CP) decomposition \cite{harshman1970foundations, carroll1970analysis, kiers2000towards} by setting the core tensor $\bm{\mathcal{G}}$ as an order-$I$ super diagonal tensor: $\bm{\mathcal{G}} \in \mathbb{R}^{M^I}$, $M=M_1=\cdots=M_I$, all zero entries except the diagonal elements. If we further set the diagonal elements of $\bm{\mathcal{G}}$ to be 1, the Tucker decomposition in Eq. \ref{eq:tucker_multi_output} becomes:

\begin{equation} \label{eq:CP_decomposition}
    \mathcal{J}\bm{u}(\bm{x}) = \sum^{M}_{m=1}{ \lbrack \mathcal{J}\prescript{m}{1}{\bm{u}}(x_1) \odot \dots \odot \mathcal{J}\prescript{m}{I}{\bm{u}}(x_I) \rbrack},
\end{equation}

where $\mathcal{J}\prescript{m}{i}{\bm{u}}(x_i) \in  \mathbb{R}^{L}$ and $\odot$ represent multiplication in elements. In CP decomposition, the core tensor is no longer trainable, thus the total trainable parameter becomes $ ML \sum^{I}_{i}{J_i} $.

It is important to note that the interpolation function $N^{(j)} (x_i)$ in Eq.\ref{eq:interpolation} is one-dimensional. In other words, both Tucker and CP decomposition replace a high-dimensional interpolation with one-dimensional interpolations that dramatically reduce the number of trainable parameters (or DoFs). As given in Table \ref{tab:INN_trainable_parameters}, the number of trainable parameters of the full interpolation scales exponentially with the discretization $J_i$, whereas the Tucker and CP decomposition scale linearly with $J_i$. Considering the trainable parameters of multi-layer perceptron (MLP) scales quadratically with the number of hidden neurons, INNs with tensor decomposition (TD) may dramatically reduce the model complexity and computing requirements.

However, the reconstructed basis using TD loses expressibility because of the separated variables. There is no cross-term in the functional space of TD. Nevertheless, we can transform this into an advantage by enriching the 1D approximations with adaptive activation (see SI Section 2.2) for the Q-step message passing. Therefore, we can find a hybrid architecture that preserves some of the advantages of numerical methods (i.e., FEM) and neural network methods while acknowledging that it loses something from both.

\begin{table}[htbp]
  \centering
  \caption{Number of trainable parameters (or degrees of freedom, DoFs) of full interpolation, Tucker decomposition, and CP decomposition for an $I$-input single output $(L=1)$ relationship. The input domain is discretized with a regular grid of $J_1 \times J_2\times \cdots \times J_I$ discretization. We assume that the nodal coordinates $(\bm{x}^{(j)} )$ are fixed.}
  \label{tab:INN_trainable_parameters}
  \begin{tabular}{ccccccc}
    \toprule
    & Full interpolation & Tucker decomposition & CP decomposition  \\
    \midrule
    Trainable parameters & $\prod^{I}_{i}{J_i}$ & $\prod^{I}_{i}{M_i} + \sum^{I}_{i}{M_i J_i} $ & $ M \sum^{I}_{i}{J_i} $ \\
    \bottomrule
  \end{tabular}
\end{table}

\subsection{Loss function of INN trainer}
\label{subsec:INN_trainer_loss}

A general INN forward propagation is provided in Eq.\ref{eq:INN_forward_propagation} as a tensor contraction. INN can be used for data training as any other neural network architecture. While MLPs optimize weights and biases during training, INN trainers optimize nodal values $\bm{U} = \{\bm{u}^{(j)} \}_{j=1,\cdots,J}$ (and, if needed, nodal coordinates ($\bm{X} = \{\bm{x}^{(j)} \}_{j=1,\cdots,J}$)) under a given loss function and training data. Consider a regression problem with $K$ labeled data: $(\bm{x}^*_k, \bm{u}^*_k)$, $k=1,\cdots,K$; and $\bm{x}^*_k \in \mathbb{R}^I, \bm{u}^*_k \in \mathbb{R}^L$. The superscript $*$ denotes the data. A mean squared error (MSE) loss function for this regression problem is defined as:

\begin{equation} \label{eq:Training_loss}
    loss(\bm{U}, \bm{X}) = \frac{1}{K} \sum_{k} (\mathcal{J}\bm{u}(\bm{x}^*_k) - \bm{u}^*_k)^2.
\end{equation}

Finally, an optimization is formulated as follows.

\begin{equation} \label{eq:Training_optimization}
    \underset{\bm{U}, \bm{X}}{\text{minimize}} \ loss(\bm{U}, \bm{X}).
\end{equation}

\subsection{Loss function of INN solver}
\label{subsec:INN_solver_loss}

An INN solver generalizes classical numerical methods such as FEM and meshfree, as well as model order reduction methods such as proper generalized decomposition (PGD) \cite{chinesta2011short} and tensor decomposition (TD) \cite{zhang2022hidenn, guo2024convolutional}. 
Here, we introduce a formulation with CP decomposition to solve the generalized space $(\bm{x})$ - time $(t)$ - parameter $(\bm{\theta})$ (S-T-P) PDE, whose computational cost is prohibitively high for most numerical methods and machine learning approaches \cite{chinesta2011short}. 

Consider a parameterized space-time PDE:

\begin{equation} \label{eq:INN-TD_solver_loss}
\begin{aligned}
    \mathcal{L}\bm{u}(\bm{x}, t, \bm{\theta}) = 0,
\end{aligned}
\end{equation}

where $\mathcal{L}$ is the general partial differential operator (can be linear or nonlinear). The INN solver uses the same neural network structure as the trainer, but the loss function varies with the equations to be solved. It is formulated as the weighted summation of the PDE residual \cite{pruliere2010deterministic,kharazmi2019variational, kharazmi2021hp}. An INN solution field in the S-T-P domain can be written as:

\begin{equation} \label{eq:INN-TD_solver}
    \begin{split}
    \mathcal{J}{\bm{u}}(\bm{x},t,\bm{\theta}; \bm{U}, \bm{X}) & = \sum_{m=1}^{M} \lbrack \mathcal{J}\prescript{m}{x_1}{\bm{u}}(x_1) \rbrack \odot \cdots \odot \lbrack \mathcal{J}\prescript{m}{x_d}{\bm{u}}(x_d) \rbrack \odot \lbrack \mathcal{J}\prescript{m}{t}{\bm{u}}(x_t) \rbrack \\ &\odot \lbrack \mathcal{J}\prescript{m}{\theta_1}{\bm{u}}(\theta_1) \rbrack \odot \cdots \odot \lbrack \mathcal{J}\prescript{m}{\theta_k}{\bm{u}}(\theta_k) \rbrack, 
    \end{split}
\end{equation}

where $d$ and $k$ are the spatial dimension and the number of parameters, respectively. Similar to the trainer, the goal is to find nodal values $\bm{U}$ (and nodal coordinates $\bm{X}$, if one wants to adapt the mesh).
As a result, the INN obtains the S-T-P solution by minimizing the loss function:

\begin{equation} \label{eq:INN-TD_STP_solver_loss}
\begin{aligned}
    \underset{\bm{U}, \bm{X}}{\text{minimize}} \int \delta \bm{u} \cdot \mathcal{L} \left[\mathcal{J}{\bm{u}}(\bm{x},t,\bm{\theta};  \bm{U}, \bm{X})\right]d\bm{x} dt d\bm{\theta},
\end{aligned}
\end{equation}
where $\delta \bm{u}$ is the weight function and can be defined using Galerkin, Petrov-Galerkin, collocation or quadratic formulation.
 Due to the Kronecker delta property of INN interpolation functions, Dirichlet boundary conditions and initial conditions can be strongly imposed. As a result, only the weighted sum residual is considered in Eq. \ref{eq:INN-TD_solver} as a loss function. This distinguishes INNs from most data-driven PDE solvers that weakly impose these conditions \cite{raissi2019physics}. See Section 4 of the SI for detailed derivations of the loss function and solution scheme. INN solvers applied to metal additive manufacturing and the linear elastic solid mechanics problem are introduced in Section \ref{subsec:applications_solving} and SI Section 4, respectively.





\subsection{Loss function of INN calibrator}
\label{subsec:INN_calibrator_loss}

The word calibration in mathematics refers to a reverse process of regression, where a known or measured observation of the output variables ($\bm{u}^*$) is used to predict the corresponding input variables ($\bm{x}^*$). By definition, it is analogous to solving an inverse problem in engineering design. To build a good calibrator, having an accurate forward model is of paramount importance, followed by building a good optimizer for the inverse problem. 
A trained or solved INN has the potential to be a superior candidate for the forward model inside a calibrator because it is fully differentiable and accurate, equipped with fast inference time ($\sim$milliseconds). A general formulation of the INN calibrator is described below:

\begin{equation} \label{eq:INN-TD_calibrator}
\begin{aligned}
    \bm{x}^* = \underset{\bm{x}}{\text{argmin}} \frac{1}{K} \sum_{k} (\mathcal{J}\bm{u}(\bm{x}) - \
    \bm{u}^*_k)^2,
\end{aligned}
\end{equation}

where $\mathcal{J}\bm{u}(\bm{x})$ is the trained or solved forward model, $\bm{u}^*_k$ is the $k$-th measured observation and $\bm{x}^*$ is the calibrated input variable. The INN calibrator applied to the heat source calibration task in metal additive manufacturing can be found in Section \ref{subsec:applications_solving} while other benchmarks can be found in SI Section 3.


\subsection{1D Poisson's equation with error estimator}
\label{subsec:Poisson_equation}

We borrow the manufactured problem first introduced in \cite{zhang2021hierarchical}. The 1D Poisson's equation is defined as:

\begin{equation} \label{eq:Poisson_eqn}
    \begin{aligned}
    \Delta u(x) + b(x) = 0, \quad x\in [0,10],  \\  u(x=0) = u(x=10) = 0 ,
    \end{aligned}
\end{equation}

where $\Delta$ is the Laplace operator. The manufactured solution becomes
\begin{equation} \label{eq:manufactured_u}
    \begin{aligned}
    u(x) = (e^{-\pi(x-2.5)^2}-e^{-6.25\pi}) + 2(e^{-\pi(x-7.5)^2}-e^{-56.25\pi}) - \frac{e^{-6.25\pi}-e^{-56.25\pi}}{10}x
    \end{aligned}
\end{equation}

when the body force is given as:
\begin{equation} \label{eq:manufactured_b}
    \begin{aligned}
    b(x) = -\frac{4\pi^2(x-2.5)^2-2\pi}{e^{\pi(x-2.5)^2}} - \frac{8\pi^2(x-7.5)^2-4\pi}{e^{\pi(x-7.5)^2}}.
    \end{aligned}
\end{equation}

The $H^1$ norm error estimator is defined as:

\begin{equation} \label{eq:H1_norm}
    \begin{aligned}
    \|e\|_{H^1} = \|u-u^h\|_{H^1} = \frac{\left[\int_{\Omega} (u-u^h)^2 \, dx + \int_{\Omega} (\frac{du}{dx}-\frac{du^h}{dx})^2 \, dx \right]^{1/2}
    }{\left[\int_{\Omega} (u)^2 \, dx + \int_{\Omega} (\frac{du}{dx})^2 \, dx\right]^{1/2}},
    \end{aligned}
\end{equation}

where $u^h(x)$ is the approximated solution.

\section*{Acknowledgments}
We express our sincere gratitude to Dr. Gino Domel, Mr. Joseph P. Leonor, Mr. Stefan Knapik, and Mr. Vispi Karkaria from Northwestern University for their invaluable comments and support throughout this work. 

\section*{Code availability}
The computer codes used for the benchmarking can be found at \href{https://github.com/hachanook/pyinn}{https://github.com/hachanook/pyinn}. Codes for additive manufacturing examples can be provided upon a reasonable request.

\bibliographystyle{elsarticle-num.bst}
\bibliography{references.bib}

\begin{thebibliography}{10}
\expandafter\ifx\csname url\endcsname\relax
  \def\url#1{\texttt{#1}}\fi
\expandafter\ifx\csname urlprefix\endcsname\relax\def\urlprefix{URL }\fi
\expandafter\ifx\csname href\endcsname\relax
  \def\href#1#2{#2} \def\path#1{#1}\fi

\bibitem{Karpathy_2017}
A.~Karpathy, \href{https://karpathy.medium.com/software-2-0-a64152b37c35}{Software 2.0} (Mar 2017).
\newline\urlprefix\url{https://karpathy.medium.com/software-2-0-a64152b37c35}

\bibitem{moor2023foundation}
M.~Moor, O.~Banerjee, Z.~S.~H. Abad, H.~M. Krumholz, J.~Leskovec, E.~J. Topol, P.~Rajpurkar, Foundation models for generalist medical artificial intelligence, Nature 616~(7956) (2023) 259--265.

\bibitem{vaswani2017attention}
A.~Vaswani, N.~Shazeer, N.~Parmar, J.~Uszkoreit, L.~Jones, A.~N. Gomez, {\L}.~Kaiser, I.~Polosukhin, Attention is all you need, Advances in neural information processing systems 30 (2017).

\bibitem{lu2021learning}
L.~Lu, P.~Jin, G.~Pang, Z.~Zhang, G.~E. Karniadakis, Learning nonlinear operators via deeponet based on the universal approximation theorem of operators, Nature machine intelligence 3~(3) (2021) 218--229.

\bibitem{zhang2022simulation}
C.~Zhang, A.~Shafieezadeh, Simulation-free reliability analysis with active learning and physics-informed neural network, Reliability Engineering \& System Safety 226 (2022) 108716.

\bibitem{goswami2022deep}
S.~Goswami, K.~Kontolati, M.~D. Shields, G.~E. Karniadakis, Deep transfer operator learning for partial differential equations under conditional shift, Nature Machine Intelligence 4~(12) (2022) 1155--1164.

\bibitem{park2023convolution}
C.~Park, Y.~Lu, S.~Saha, T.~Xue, J.~Guo, S.~Mojumder, D.~W. Apley, G.~J. Wagner, W.~K. Liu, Convolution hierarchical deep-learning neural network (c-hidenn) with graphics processing unit (gpu) acceleration, Computational Mechanics (2023) 1--27.

\bibitem{grossmann2024can}
T.~G. Grossmann, U.~J. Komorowska, J.~Latz, C.-B. Sch{\"o}nlieb, Can physics-informed neural networks beat the finite element method?, IMA Journal of Applied Mathematics (2024) hxae011.

\bibitem{mcgreivy2024weak}
N.~McGreivy, A.~Hakim, Weak baselines and reporting biases lead to overoptimism in machine learning for fluid-related partial differential equations, Nature Machine Intelligence (2024) 1--14.

\bibitem{amin2024npj}
A.~A. Amin, Y.~Li, Y.~Lu, X.~Xie, Z.~Gan, S.~Mojumder, G.~J. Wagner, W.~K. Liu, Physics guided heat source for quantitative prediction of in718 laser additive manufacturing processes, npj Computational Materials 10~(37) (2024).

\bibitem{leonor2024go}
J.~P. Leonor, G.~J. Wagner, Go-melt: Gpu-optimized multilevel execution of lpbf thermal simulations, Computer Methods in Applied Mechanics and Engineering 426 (2024) 116977.

\bibitem{shih2021fe}
M.~Shih, K.~Chen, T.~Lee, D.~Tarng, C.~Hung, Fe simulation model for warpage evaluation of glass interposer substrate packages, IEEE Transactions on Components, Packaging and Manufacturing Technology 11~(4) (2021) 690--696.

\bibitem{liu2022eighty}
W.~K. Liu, S.~Li, H.~S. Park, Eighty years of the finite element method: Birth, evolution, and future, Archives of Computational Methods in Engineering 29~(6) (2022) 4431--4453.

\bibitem{zhang2021hierarchical}
L.~Zhang, L.~Cheng, H.~Li, J.~Gao, C.~Yu, R.~Domel, Y.~Yang, S.~Tang, W.~K. Liu, Hierarchical deep-learning neural networks: finite elements and beyond, Computational Mechanics 67 (2021) 207--230.

\bibitem{saha2021hierarchical}
S.~Saha, Z.~Gan, L.~Cheng, J.~Gao, O.~L. Kafka, X.~Xie, H.~Li, M.~Tajdari, H.~A. Kim, W.~K. Liu, Hierarchical deep learning neural network (hidenn): An artificial intelligence (ai) framework for computational science and engineering, Computer Methods in Applied Mechanics and Engineering 373 (2021) 113452.

\bibitem{wu2020comprehensive}
Z.~Wu, S.~Pan, F.~Chen, G.~Long, C.~Zhang, S.~Y. Philip, A comprehensive survey on graph neural networks, IEEE transactions on neural networks and learning systems 32~(1) (2020) 4--24.

\bibitem{zhou2020graph}
J.~Zhou, G.~Cui, S.~Hu, Z.~Zhang, C.~Yang, Z.~Liu, L.~Wang, C.~Li, M.~Sun, Graph neural networks: A review of methods and applications, AI open 1 (2020) 57--81.

\bibitem{ju2024comprehensive}
W.~Ju, Z.~Fang, Y.~Gu, Z.~Liu, Q.~Long, Z.~Qiao, Y.~Qin, J.~Shen, F.~Sun, Z.~Xiao, et~al., A comprehensive survey on deep graph representation learning, Neural Networks (2024) 106207.

\bibitem{kolda2009tensor}
T.~G. Kolda, B.~W. Bader, Tensor decompositions and applications, SIAM review 51~(3) (2009) 455--500.

\bibitem{zhang2022hidenn}
L.~Zhang, Y.~Lu, S.~Tang, W.~K. Liu, Hidenn-td: Reduced-order hierarchical deep learning neural networks, Computer Methods in Applied Mechanics and Engineering 389 (2022) 114414.

\bibitem{lu2023convolution}
Y.~Lu, H.~Li, L.~Zhang, C.~Park, S.~Mojumder, S.~Knapik, Z.~Sang, S.~Tang, D.~W. Apley, G.~J. Wagner, et~al., Convolution hierarchical deep-learning neural networks (c-hidenn): finite elements, isogeometric analysis, tensor decomposition, and beyond, Computational Mechanics (2023) 1--30.

\bibitem{li2023convolution}
H.~Li, S.~Knapik, Y.~Li, C.~Park, J.~Guo, S.~Mojumder, Y.~Lu, W.~Chen, D.~W. Apley, W.~K. Liu, Convolution hierarchical deep-learning neural network tensor decomposition (c-hidenn-td) for high-resolution topology optimization, Computational Mechanics (2023) 1--20.

\bibitem{attar2014selective}
H.~Attar, M.~B{\"o}nisch, M.~Calin, L.-C. Zhang, S.~Scudino, J.~Eckert, Selective laser melting of in situ titanium--titanium boride composites: Processing, microstructure and mechanical properties, Acta Materialia 76 (2014) 13--22.

\bibitem{liao2023efficient}
S.~Liao, A.~Golgoon, M.~Mozaffar, J.~Cao, Efficient gpu-accelerated thermomechanical solver for residual stress prediction in additive manufacturing, Computational Mechanics 71~(5) (2023) 879--893.

\bibitem{alet2019graph}
F.~Alet, A.~K. Jeewajee, M.~B. Villalonga, A.~Rodriguez, T.~Lozano-Perez, L.~Kaelbling, Graph element networks: adaptive, structured computation and memory, in: International Conference on Machine Learning, PMLR, 2019, pp. 212--222.

\bibitem{michelis2022physics}
M.~Y. Michelis, R.~K. Katzschmann, Physics-constrained unsupervised learning of partial differential equations using meshes, arXiv preprint arXiv:2203.16628 (2022).

\bibitem{gao2022physics}
H.~Gao, M.~J. Zahr, J.-X. Wang, Physics-informed graph neural galerkin networks: A unified framework for solving pde-governed forward and inverse problems, Computer Methods in Applied Mechanics and Engineering 390 (2022) 114502.

\bibitem{xue2022physics}
T.~Xue, Z.~Gan, S.~Liao, J.~Cao, Physics-embedded graph network for accelerating phase-field simulation of microstructure evolution in additive manufacturing, npj Computational Materials 8~(1) (2022) 201.

\bibitem{tian2013extra}
R.~Tian, Extra-dof-free and linearly independent enrichments in gfem, Computer Methods in Applied Mechanics and Engineering 266 (2013) 1--22.

\bibitem{liu1995reproducing}
W.~K. Liu, S.~Jun, Y.~F. Zhang, Reproducing kernel particle methods, International journal for numerical methods in fluids 20~(8-9) (1995) 1081--1106.

\bibitem{liu1995reproducing2}
W.~K. Liu, S.~Jun, S.~Li, J.~Adee, T.~Belytschko, Reproducing kernel particle methods for structural dynamics, International Journal for Numerical Methods in Engineering 38~(10) (1995) 1655--1679.

\bibitem{leng2019super}
Y.~Leng, X.~Tian, J.~T. Foster, Super-convergence of reproducing kernel approximation, Computer Methods in Applied Mechanics and Engineering 352 (2019) 488--507.

\bibitem{tucker1963implications}
L.~R. Tucker, Implications of factor analysis of three-way matrices for measurement of change, Problems in measuring change 15~(122-137) (1963) 3.

\bibitem{tucker1966some}
L.~R. Tucker, Some mathematical notes on three-mode factor analysis, Psychometrika 31~(3) (1966) 279--311.

\bibitem{harshman1970foundations}
R.~A. Harshman, et~al., Foundations of the parafac procedure: Models and conditions for an “explanatory” multi-modal factor analysis, UCLA working papers in phonetics 16~(1) (1970) 84.

\bibitem{carroll1970analysis}
J.~D. Carroll, J.-J. Chang, Analysis of individual differences in multidimensional scaling via an n-way generalization of “eckart-young” decomposition, Psychometrika 35~(3) (1970) 283--319.

\bibitem{kiers2000towards}
H.~A. Kiers, Towards a standardized notation and terminology in multiway analysis, Journal of Chemometrics: A Journal of the Chemometrics Society 14~(3) (2000) 105--122.

\bibitem{jax2018github}
J.~Bradbury, R.~Frostig, P.~Hawkins, M.~J. Johnson, C.~Leary, D.~Maclaurin, G.~Necula, A.~Paszke, J.~Vander{P}las, S.~Wanderman-{M}ilne, Q.~Zhang, \href{http://github.com/jax-ml/jax}{{JAX}: composable transformations of {P}ython+{N}um{P}y programs} (2018).
\newline\urlprefix\url{http://github.com/jax-ml/jax}

\bibitem{surjanovic2013virtual}
S.~Surjanovic, D.~Bingham, Virtual library of simulation experiments: test functions and datasets, Simon Fraser University, Burnaby, BC, Canada, accessed May 13 (2013) 2015.

\bibitem{shin2020convergence}
Y.~Shin, J.~Darbon, G.~E. Karniadakis, On the convergence of physics informed neural networks for linear second-order elliptic and parabolic type pdes, arXiv preprint arXiv:2004.01806 (2020).

\bibitem{bhavar2017review}
V.~Bhavar, P.~Kattire, V.~Patil, S.~Khot, K.~Gujar, R.~Singh, A review on powder bed fusion technology of metal additive manufacturing, Additive manufacturing handbook (2017) 251--253.

\bibitem{cai2023review}
Y.~Cai, J.~Xiong, H.~Chen, G.~Zhang, A review of in-situ monitoring and process control system in metal-based laser additive manufacturing, Journal of Manufacturing Systems 70 (2023) 309--326.

\bibitem{kozjek2022data}
D.~Kozjek, F.~M. Carter~III, C.~Porter, J.-E. Mogonye, K.~Ehmann, J.~Cao, Data-driven prediction of next-layer melt pool temperatures in laser powder bed fusion based on co-axial high-resolution planck thermometry measurements, Journal of Manufacturing Processes 79 (2022) 81--90.

\bibitem{li2024statistical}
Y.~Li, S.~Mojumder, Y.~Lu, A.~Al~Amin, J.~Guo, X.~Xie, W.~Chen, G.~J. Wagner, J.~Cao, W.~K. Liu, Statistical parameterized physics-based machine learning digital shadow models for laser powder bed fusion process, Additive Manufacturing 87 (2024) 104214.

\bibitem{yeung2020residual}
H.~Yeung, B.~Lane, A residual heat compensation based scan strategy for powder bed fusion additive manufacturing, Manufacturing letters 25 (2020) 56--59.

\bibitem{karkaria2024towards}
V.~Karkaria, A.~Goeckner, R.~Zha, J.~Chen, J.~Zhang, Q.~Zhu, J.~Cao, R.~X. Gao, W.~Chen, Towards a digital twin framework in additive manufacturing: Machine learning and bayesian optimization for time series process optimization, Journal of Manufacturing Systems (2024).

\bibitem{murdoch2019definitions}
W.~J. Murdoch, C.~Singh, K.~Kumbier, R.~Abbasi-Asl, B.~Yu, Definitions, methods, and applications in interpretable machine learning, Proceedings of the National Academy of Sciences 116~(44) (2019) 22071--22080.

\bibitem{liu2016self}
Z.~Liu, M.~Bessa, W.~K. Liu, Self-consistent clustering analysis: an efficient multi-scale scheme for inelastic heterogeneous materials, Computer Methods in Applied Mechanics and Engineering 306 (2016) 319--341.

\bibitem{yu2019self}
C.~Yu, O.~L. Kafka, W.~K. Liu, Self-consistent clustering analysis for multiscale modeling at finite strains, Computer Methods in Applied Mechanics and Engineering 349 (2019) 339--359.

\bibitem{chinesta2011short}
F.~Chinesta, P.~Ladeveze, E.~Cueto, A short review on model order reduction based on proper generalized decomposition, Archives of Computational Methods in Engineering 18~(4) (2011) 395--404.

\bibitem{guo2024convolutional}
J.~Guo, C.~Park, X.~Xie, Z.~Sang, G.~J. Wagner, W.~K. Liu, Convolutional hierarchical deep learning neural networks-tensor decomposition (c-hidenn-td): a scalable surrogate modeling approach for large-scale physical systems, arXiv preprint arXiv:2409.00329 (2024).

\bibitem{pruliere2010deterministic}
E.~Pruliere, F.~Chinesta, A.~Ammar, On the deterministic solution of multidimensional parametric models using the proper generalized decomposition, Mathematics and Computers in Simulation 81~(4) (2010) 791--810.

\bibitem{kharazmi2019variational}
E.~Kharazmi, Z.~Zhang, G.~E. Karniadakis, Variational physics-informed neural networks for solving partial differential equations, arXiv preprint arXiv:1912.00873 (2019).

\bibitem{kharazmi2021hp}
E.~Kharazmi, Z.~Zhang, G.~E. Karniadakis, hp-vpinns: Variational physics-informed neural networks with domain decomposition, Computer Methods in Applied Mechanics and Engineering 374 (2021) 113547.

\bibitem{raissi2019physics}
M.~Raissi, P.~Perdikaris, G.~E. Karniadakis, Physics-informed neural networks: A deep learning framework for solving forward and inverse problems involving nonlinear partial differential equations, Journal of Computational physics 378 (2019) 686--707.

\end{thebibliography}

\end{document}